\documentclass{article} 
\usepackage{iclr2020_conference,times}

\usepackage{amsmath,amsfonts,bm}

\def\eqref#1{equation~\ref{#1}}

\def\1{\bm{1}}

\DeclareMathAlphabet{\mathsfit}{\encodingdefault}{\sfdefault}{m}{sl}
\SetMathAlphabet{\mathsfit}{bold}{\encodingdefault}{\sfdefault}{bx}{n}

\usepackage{microtype}
\usepackage{graphicx}
\usepackage{amsmath,amssymb}
\usepackage{subfigure}
\usepackage{csquotes}
\usepackage{multirow}
\usepackage{booktabs} 
\usepackage{xcolor}
\usepackage{multibib}
\usepackage{dirtytalk}
\newcites{appen}{Reference}

\title{Optimizing Medical Treatment for Sepsis in\\ Intensive Care: from Reinforcement Learning to Pre-Trial Evaluation}

\author{Luchen Li$^1$, Ignacio Albert-Smet$^2$ \& Aldo A. Faisal$^{1,2,3,4}$ \\
$^1$Brain \& Behaviour Lab: Dept. of Computing, $^2$Dept. of of Bioengineering,\\$^3$ Behaviour Analytics Lab, Data Science Institute,\\$^4$ UKRI Centre for Doctoral Training in AI for Healthcare
\\
Imperial College London, London, UK \\
\texttt{l.li17@imperial.ac.uk,a.faisal@imperial.ac.uk} \\
}

\iclrfinalcopy 
\begin{document}

\maketitle

\begin{abstract}
Our aim is to establish a framework where reinforcement learning (RL) of optimizing interventions retrospectively allows us a regulatory compliant pathway to prospective clinical testing of the learned policies in a clinical deployment. We focus on infections in intensive care units which are one of the major causes of death and difficult to treat because of the complex and opaque patient dynamics, and the clinically debated, highly-divergent set of intervention  policies required by each individual patient, yet intensive care units are naturally data rich.  In our work, we build on RL approaches in healthcare (\say{AI Clinicians}), and learn off-policy continuous dosing policy of pharmaceuticals for sepsis treatment using  historical intensive care data under partially observable MDPs (POMDPs).
POMPDs capture   uncertainty in patient  state better by taking in all historical information, yielding an efficient representation, which we investigate through ablations. We compensate for the lack of exploration in our retrospective data by evaluating each encountered state with a best-first tree search. We mitigate state distributional shift by optimizing our policy in the vicinity of the clinicians' compound policy. Crucially, we evaluate our model recommendations using not only conventional policy evaluations but a novel framework that incorporates human experts: a model-agnostic pre-clinical evaluation method to estimate the accuracy and uncertainty of clinician's decisions versus our system recommendations when confronted with the same individual patient history (\say{shadow mode}). 
\end{abstract}

\section{Introduction}
Automatic treatment optimization based on reinforcement learning (RL) has been explored in simulated patients \cite{10Ernst2006CDB, 11Bothe2013TheUO, Lowery2013}.
The dynamic interactions between patients and clinicians, how patient state and clinical actions interact, have been 
successfully modelled with reinforcement learning (RL) as the mathematical framework of a Markov Decision Process (MDP), as established by \cite{Komorowski2018naturemed} and applied to sepsis. Sepsis is at a global scale (\cite{Rudd2020}) a major burden on healthcare systems is ubiquitous. Because of the easily scalable nature of software, such recommender systems have the potential to become an affordable means to improve patient care. Thus, while the application to intensive care and sepsis are specific, they illustrate what is generally 
hard in reinforcement learning for clinical interventions 
\cite{Gottesman2019}. 

At each time step, the patient's 
physiological state $s_t$ gives rise to the clinician's 
intervention decision $a_t$. Upon providing an immediate 
feedback $r_t = r(s_t, a_t)$, the patient transits to the 
next state $s_{t+1}$. However, the patient states may not 
necessarily be readily observable as a result of an 
incomplete or inaccurate selection of medical measurements to base decisions on, noisy input due to human error or 
apparatus limitations, and the inconsistent frequency with 
which measurements are taken. A less restricted framework 
to  clinical scenarios is the Partially-Observable Markov 
Decision Process (POMDP) by \cite{30Sondik1978TOCP}. A 
POMDP can always be transformed to an MDP given that a 
Markovian  history-dependent representation of the 
environment is found (see Appendix~\ref{section:supp_related_work}).
We use a variational generative model to  predict the observation at the next step given the history of observation and actions. We implement history dependency and memory through a recurrent neural network (RNN). The training of the RNN is embedded as an auxiliary task of learning. Our modular approach decouples state representation learning from policy learning and enables to evaluate this independently (e.g. ablation) and interrogate it for explainability. Subsequently, we learn a continuous policy over vasopressors and intravenous fluids by iteratively training an actor-critic model and conducting an improvable best-first tree search to explore potentially more valuable states.

In addition to conventional policy evaluations, we collect 
data from clinicians to compare with the models. We propose a model-agnostic pre-clinical validation method that follows the guidelines proposed by \cite{Gottesman2019} and can serve as 
a benchmark for comparing the actions of different actors, 
both human and artificial. The metrics obtained can be used 
to check the proposed actions in a safe and risk-conscious 
manner. For training data and pipeline please refer to our online repository http://www.imperial.ac.uk/artificial-intelligence/research/healthcare/ai-clinician/.





\section{Treatment Optimization}
\subsection{State representation}

In a POMDP, the whole history trace of observations and actions $\{o_0, a_0, \dots, o_{t-1}, a_{t-1}, o_t\}$ is used to infer the action $a_t$ at time step $t$. We summarize this history information through an RNN, with input $\{a_{t-1}, o_t\}$, and use the RNN hidden state as the representation of the patient's physiological state $s_t$.

We argue that the state representation can be learned in a more supervised way than as a feature encoder. This way, environmental information can be grasped accurately even before the policy has been updated enough to make good decisions. Specifically, we implement the conditional variational auto-encoder (CVAE) from \cite{vaekingma14} to represent the state in a way that, in conjunction with an independent action, can most probably predict the next observation:
\begin{equation}
\mathop{\mathrm{max}}_{\mathrm{RNN}}~p(o_{t+1}|s_t, a_t)
\end{equation}



\subsection{Policy optimization}
Theoretically learning a policy $\pi$ from patient trajectories generated by the clinicians' behavior policy $\pi_b$ can be conferred by any off-policy RL approach.
We optimize $\pi$ via actor-critic, with importance sampling (IS) to reweigh the impact of an action (\cite{9Sutton98a}). In the meantime, to explore potentially better trajectories (i.e. of states with higher values) than those already generated, we execute a best-first tree search for each encountered state to explore its state value beyond the limitation of either $\pi_b$ or $\pi$. In a nutshell, we iteratively train a global actor-critic model and carry out a local heuristic tree search for the current state. The actor-critic model provides leaf-node estimators required for expanding and backing-up the tree, which in turn is updated by the backed-up value at the root. The actor-critic not only improves the tree search over time, but also exempts tree search during the inference phase with a trained global policy.

For implementation details of deep RL please see Appendix~\ref{section:supp_implementation}. The state distributional shift due to no further interactions with the environment is discussed and mitigated in Appendix~\ref{section:supp_distr_shift}.

\subsubsection{Global actor-critic}
\label{section:actor_critic}
We estimate the off-policy advantage function with V-trace (\cite{Espeholt2018IMPALA}), with the distinction that we take into account all the subsequent trace.
Similar to the up-going return in \cite{starcraf19}, we devise an \textit{up-going advantage} that considers only the immediate TD error when farther steps result in a negative advantage. Our advantage function is thereby estimated backwards as:
\begin{equation}
A(s_t, a_t) = \begin{cases}
\delta_t, ~~~~~~~~~~~~~~~~~~~~~~~~~~~~~~~~~~~~~~~~~~~~~~~~~~~~\mathrm{last~ nonterminal~ step}\\
\gamma c_t\max\big(A(s_{t+1}, a_{t+1}), ~0\big)+\delta_t, ~~\mathrm{otherwise}
\end{cases}
\end{equation}
where $\delta_t$ is the off-policy temporal difference error for $V$ at time $t$, $c_t$ is a contracted IS ratio. The state value function is regressed onto an estimation target returned by the tree search, by minimizing the conventional mean squared error loss.

\subsubsection{Local heuristic tree search}

In order for the tree search\footnote{In this subsection, we will omit the subscript denoting time step, as the root is the only node in a tree that is a virtually encountered state. Superscript is used to denote depth in the tree.} to improve more efficiently upon the global estimators, we make additional use of the CVAE to grow the tree and back up from the most likely reachable regions, by selecting from the leaf nodes $\mathcal{F}$ to expand:
\begin{equation}\label{eq:node2expand}
s^* = \mathop{\mathrm{arg\,max}}_{s \in\mathcal{F}}~\gamma^{D(s)-1} \prod_{d=0}^{D(s)-1} \Tilde{p}\big(s^{d+1}|s^d, a_\mathcal{T}(s^d)\big) \delta_{\mathcal{T}(s^0, s)}(s^d)
\end{equation}
where $D(s)$ denotes the depth of node $s$, $\delta_x(\cdot)$ is a Dirac delta function concentrated at $x$, and $\delta_{\mathcal{T}(s^0, s)}(\cdot)$ is used to represent a node within the subtree from root $s^0$ to leaf $s$. $a_\mathcal{T}(s^d)$ denotes the deterministic tree policy in $s^d$ and will be discussed further. A fixed amount of direct descendants $o^{d+1}$'s are simulated from $s^d$ by the CVAE, each afterwards being sent to the RNN to deterministically yield a $s^{d+1}$. $\Tilde{p}(s^{d+1}|s^d, a_\mathcal{T}(s^d))$ is the relative probability of reaching the child node $s^{d+1}$ from its parents, normalized across all siblings.

At each expansion, we choose among the leaf nodes the most likely reachable from the root by \eqref{eq:node2expand}, whose value thereby accounts for the most in the value of the root. The best-first characteristic of our tree search is two-fold, the second being a deterministic tree policy that always selects the optimal action with respect to the current state value function estimator to expand a node
\begin{equation}\label{eq:tree_policy_det}
a_\mathcal{T}(s^d) = \mathop{\mathrm{arg\,max}}_a~ r(s^d, a) + \gamma\sum_{s^{d+1}\in C(s^d, a)} \Tilde{p}(s^{d+1}|s^d, a)V(s^{d+1})
\end{equation}
with $C(s, a)$ representing the set of child nodes of $s$ by taking $a$. A softmax version of \eqref{eq:tree_policy_det} can also be implemented, but we observe unremarkable improvement to the deterministic policy.

The state values of all tree nodes are back-propagated in a bottom-up fashion via Bellman update \cite{52Ross2007AAO} to the root after a small and fixed number of expansions, with the values of the leafs provided by the current state value function approximator $V(s)$
\begin{align}
v_{\mathcal{T}}(s^d) &= \begin{cases}
V(s^d), ~~~~~~~~~~~~~~~~~~~~~~~~~~~~~~~~~~~~~~~~~~~~~~~~~~~~~~~~~~~~~~~~~~~~~~~~~~~~~~~~~~~~~~~~~~s^d\mathrm{~is~a~leaf ~ node}\\
r(s^d, a_\mathcal{T}(s^d)) + \gamma\sum_{s^{d+1}\in C(s^d, a_\mathcal{T}(s^d))} \Tilde{p}\big(s^{d+1}|s^d, a_\mathcal{T}(s^d)\big)v_{\mathcal{T}}(s^{d+1}), ~\mathrm{otherwise}
\end{cases}
\end{align}

$V(s^0)$ is subsequently regressed onto $v_{\mathcal{T}}(s^0)$.

\section{Preclinical Validation}
\subsection{Experimental Setup}
The aim of preclinical validation is to look at how the algorithms' proposed actions could perform in a clinical setting. To achieve this, we compare the algorithms' recommendations against actions taken by human clinicians who were given the same patient information.

A user interface (UI) was built to collect experimental data from the actions of clinicians. The UI presents the subject past patient data similar to that used to train the models, and then asks him/her to input a treatment dose of fluids and vasopressors. These inputs are in the form of a mean and a variance to help capture the uncertainty of these clinical decisions. This matches the actions of our POMDP agent, and can also be used to validate actions from discrete-action models, such as that proposed by \cite{Komorowski2018naturemed}.

The patient data to be displayed in the UI for this study is randomly selected from the test set. To produce meaningful results, there are two selection criteria. Firstly, at least half of the patients selected must have been on vasopressors at some point in their trajectory. This is done because only 35\% of the patients in the dataset receive vasopressors, which could result in few data points with which to validate these actions. The second selection criterion is length of stay, which must be long enough to create a meaningful trajectory from these points. The selected patient records had to be at least 48 hours long.

\subsection{Data Processing}

Details about the patient cohort and data pre-processing can be found in Appendix~\ref{section:supp_cohort}. After training the policy, two mathematically-defined scores (or metrics) are used to capture the relationship between what the participants recommended, what the AI agents suggested, and what the original clinician did. The Probabilistic Score (P-score) $P_{score} = p\big(a| D\big)~/~N ~~~~~ \mathrm{where} ~~~ D(a) = \sum^{N}_{i=1} \mathcal{N} \big(a; \mu_{i} , \Sigma_{i} \big)$,
is a continuous metric that defines the probability of choosing a certain action. A higher P-score indicates that the action was more likely chosen by the participant.

Meanwhile, the Count Score (C-score) is a frequency metric that accounts for the proportion of clinicians that could have chosen that action with 99\% confidence. A C-score value close to 1 means that, for most clinicians, the action is deemed as acceptable within their own uncertainty limits.

\begin{equation}
C_{score} = \sum^{N}_{i=1} c_{i}(a)~/~N ~~~~~ \mathrm{where} ~~~ c_{i}(a) =
\begin{cases}
1 ~~~ \mbox{if} ~~~ p\big(a| \mu_{i},\Sigma_{i}\big) ~~ \geq ~ 0.01 \\ \\
0 ~~~ \mbox{if} ~~~ p\big(a| \mu_{i},\Sigma_{i}\big) ~ < ~ 0.01
\end{cases}
\end{equation}

\begin{figure}
\includegraphics[width=1\textwidth]{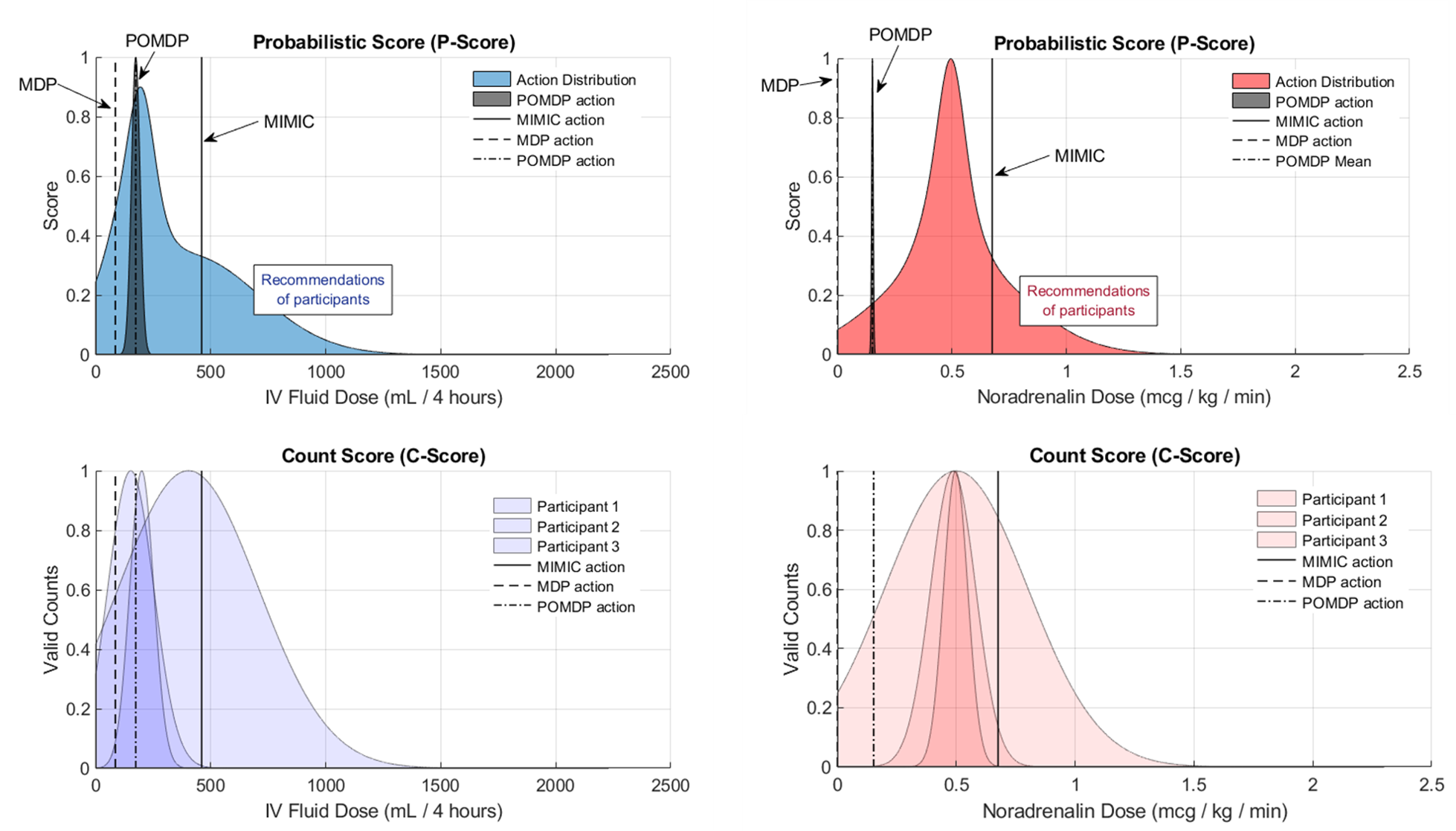}
\vspace{-0.2cm}
\caption{Example of P-scores (\emph{top}) and C-scores (\emph{bottom}) from three recommended doses of intravenous fluids (\emph{in blue, left}) and vasopressors (\emph{in red, right}). The P-score is the integral of an action with the probability density function of all recommendations. The C-score counts intersections (with 99\% confidence) of actions with individual recommendations. On the legend: \emph{MIMIC} stands for the action in the patient record, \emph{MDP} stands for the action of the discrete model from \cite{Komorowski2018naturemed}, and \emph{POMDP} stands for the model proposed in this work.}
\label{fig:PCscores_example}
\end{figure}

\section{Results}

This section presents the results obtained for our model when compared to decisions made by intensivists. For off-policy evaluations and ablation studies that do not depend on further clinical knowledge, please refer to Appendix~\ref{section:supp_OPE}.

By the time of this submission, we conducted a small pilot study to test the validation pipeline (see Table \ref{tab:PCscores}). The three participants recruited are intensive care consultants from the Imperial College Healthcare Trust with experience in sepsis treatment. We are aware that a comprehensive validation of the agents would require more subjects, and we are currently in the process of building a larger dataset.

To be trusted by the clinician, the actions recommended by the decision-support AI need to be similar enough to be considered safe and reasonable, but must also be distinct enough to improve patient care. Bear in mind that the goal is to use RL to solve a treatment optimization problem, and not a predictive or diagnostic one.

In that sense, the P-score measures how much clinicians agree with the chosen action, while the C-score is a better measure for what they find is clinically acceptable within their uncertainty. For instance, two actions that are within a reasonable dose range could have equal C-scores and very different P-scores. Compared to the P-score, the C-score allows for a certain level of disagreement and thus respects, to some degree, the agents’ \emph{exploration}. That said, the P-score of the original action (shown as \emph{MIMIC} in Table \ref{tab:PCscores}) should be similar to those proposed by the subjects. This would indicate that the data used to train the agents and shown to the clinicians in the UI sufficiently informs medical decisions in this clinical setting.

Finally, we calculated the proportion of evaluated points that received a C-score equal to zero (shown as \emph{Zero Count} in Table \ref{tab:PCscores}). These actions are outside the uncertainty thresholds of all three subjects. When a zero count is recorded for the original action, this could indicate that something happened beyond of the scope of our data (e.g.: an intervention not reflected in the monitoring data). 

\begin{table}[t]
\label{sample-table}
\begin{center}
\caption{This table contains the average validation scores observed when three separate intensive care consultants evaluated 10 patient trajectories from the MIMIC dataset. The best score for each row is shown in bold. The column labels stand for: the original action recorded (MIMIC), a discrete and non-time dependant model (MDP) by \cite{Komorowski2018naturemed}, and the model we propose in this work (POMDP). \\}
\label{tab:PCscores}
\begin{tabular}{ccccc}
\multicolumn{1}{c}{\bf ACTION} &\multicolumn{1}{c}{\bf SCORE}
&\multicolumn{1}{c}{\bf MIMIC}  &\multicolumn{1}{c}{\bf MDP} &\multicolumn{1}{c}{\bf POMDP}
\\ \hline
\multirow{3}{6em}{ IV Fluids}
     &P-Score &\bf{0.454} &0.434 &0.448
\\   &C-Score &0.578 &0.622 &\bf{0.644}
\\   &Zero Count &0.133 &0.100 &\bf{0.033}
\\ \hline

\multirow{3}{6em}{Vasopressors}
     &P-Score &\bf{0.584} &0.286 &0.488
\\   &C-Score &\bf{0.700} &0.467 &0.644 
\\   &Zero Count &\bf{0.033} &0.233 &\bf{0.033}
\\  \hline
\end{tabular}
\end{center}
\end{table}

\section{Conclusion}
We propose and evaluate an off-policy reinforcement learning method that leverages all historical information to learn fluid and vasopressor prescription for sepsis patients in intensive care, trained and tested on real patient data. In particular, the exploitation of a generative model in our work allows for accurate state representation decoupled of policy learning, and the combination of a global actor-critic model and a local tree search enables exploration into unvisited states which is intrinsically unfeasible in batch RL tasks.

In light of initial justification of our algorithmic structure in terms of a series of off-policy evaluations and ablation studies (see Appendix~\ref{section:supp_OPE}), we implement a framework for preclinical validation that is model-agnostic and can be used to compare in a safe and risk-conscious manner the actions of human and artificial clinicians.
Although additional data for a more comprehensive analysis is desired, our evaluation framework bridges the gap between an artificial clinician and a real patient by a further step.

{

}

\newpage
\appendix
\setcounter{figure}{0}

\section{Related Work}
\label{section:supp_related_work}
Automatic treatment optimization based on reinforcement learning (RL) has been explored in simulated patients \citeappen{10Ernst2006CDB, 11Bothe2013TheUO, Lowery2013}. Public-available electronic healthcare records allow medical sequential decisions to be learned from real-world experiences \citeappen{8Shortreed2011ISCD, 19Asoh2013AAI, 25Lizotte2016MOMD, 12Prasad2017MVI}, even achieving human-comparable performance \citeappen{Komorowski2018naturemed}.

These works are built on MDPs, whereas algorithms in POMDPs suffer notoriously from computational complexity because the history space grows exponentially in planning horizon. \citeappen{Hauskrecht2000, Tsoukalas2015, 24Nemati2016OptimalMD} build POMDP frameworks in medical context but nevertheless assume discrete observation and/or action spaces and impute trivial structures on environment dynamics.

In conjunction with deep learning, we learn a continuous history-dependent representation of patient state in a supervised way, and then optimize a continuous treatment policy by combining global estimators and local search. Our model is trained using real intensive care data from MIMIC-III \citeappen{Johnson2017MAF, Komorowski2018naturemed}.

\section{Learning Implementation Details}
\label{section:supp_implementation}
All actions $a$ and observations $o$ are encoded by two-layer perceptrons before passing into a network. Multiple inputs are concatenated as one vector. Both behavior policy $\pi_b$ and learned policy $\pi$ are continuous. $\pi_b$ is estimated from (another) CVAE to capture the potential multimodal-ness in clinicians' decisions.

All covariance matrices are diagonal. For the encoder in both CVAEs, the standard deviation is fixed. The log standard deviations of both CVAE decoders and of $\pi$ are dependent on the network's input. The actor-critic model is aapproximated with a two-layer perceptron followed by three separate heads: the vector Gaussian mean and variance, and the scalar state value function.

The RNN is a gated recurrent unit (GRU). All hidden layers are activated by rectified linear units (ReLUs). All network weights are initialized using orthogonal initializer. Batch normalization is not implemented.

We use RMSProp as the optimizer as it works well with RNNs. Momentum is not considered. The decay factor in RMSProp is $\alpha=0.99$. Gradients are truncated at the norm of $0.5$. The relative weights for state value approximator is $0.5$. Entropy of the learned policy is not intentionally encouraged.

Our hyperparameters under tuning include learning rate $([1e-5, 5e-4]\textnormal{~log uniform})$ and regularization factor $\varepsilon$ in RMSProp $([1e-5, 1e-1]\textnormal{~log uniform})$.

State representation learning, supervised behavior policy estimation, and reinforcement learning are carried out sequentially, though they can be undertaken simultaneously with the same respective objective.

\section{Cohort}
\label{section:supp_cohort}
\subsection{Cohort selection}
Our cohort is a retrospective database, Medical Information Mart for Intensive Care Clinical Database (MIMIC-III) \citeappen{Johnson2017MAF} that contains de-identified health-related records of patients during their stays in a hospital ICU. We include data based on the same conditions as \citeappen{Komorowski2018naturemed}, selecting adult patients in the MIMIC-III database who conform to the international consensus sepsis-3 criteria \citeappen{Singer2016}, excluding admissions in which treatment was withdrawn or mortality was not documented. For all patients, data are extracted from up to $24$ hours preceding, and until up to $48$ hours ensuing the estimated onset of sepsis, to which the time series are aligned. Time series are temporally discretized with $1$-hour intervals, resulting in $20,587$ ICU admissions or $984,010$ steps in total. $2,000$ admissions are randomly sampled and excluded from training for off-policy policy evaluation. 

The treatments and patient variables to consider are also identical to those in \citeappen{Komorowski2018naturemed}, recounted here for completeness, with \textit{value continuities} retained. Our action space consists of the maximum dose of vasoppressors administered and the total volume of intravenous fluids injected over each hourly period. The vasopressors include norepinephrine, epinephrine, vasopressin, dopamine and phenylephrine, and are converted to norepinephrine-equivalent with dose correspondence \citeappen{Brown2013vasopressor}. The intravenous fluids include boluses and background infusions of crystalloids, colloids and blood products, normalized by tonicity.

The single goal in the RL task is survival. Transitions to the terminal discharge from ICU without deceasing in $90$ days are rewarded by $+10$, those to the terminal death, either in-hospital or $90$ days from discharge, are penalized by $-10$. No intermediate reward is set so as to encourage learning from scratch.

\subsection{Data preprocessing}
The patient features that make up our observation space include $59$ continuous variables and $4$ binary ones that consist of bedside measurements, laboratory results and demographics. Both continuous features and actions are histogram equalized before input into neural networks. Missing data values are interpolated via sample-and-hold which conforms to how clinicians would conceive the situation.

\section{Distributional Shift}
\label{section:supp_distr_shift}
As with all batch reinforcement learnings, no further interaction with the environment is permitted during training. In addition to off-policy-ness which is addressed in Section~\ref{section:actor_critic}, batch RL is known to suffer from state distributional shift \citeappen{dagger11, Liu2018} that stemmed from the correspondence between a policy and the stationary state distribution. This phenomenon is detrimental as the impact of any action in regions of the state space unexplored by the behavior policy is counterfactual. Two branches of approaches are devised to alleviate this problem: 1) applying state distribution correction during gradient computation \citeappen{Johansson16, Shalit17, stateliu19}, and 2) constraining the learned policy largely to the behavior policy \citeappen{Fujimoto19, Laroche19} so that the policy induced stationary state distribution $d^\pi(s)$ can be deemed equivalent to $d^{\pi_b}(s)$.

For clinical safety, we encourage the induced states not to deviate much from those explored by clinicians, and thus include an objective of behavior cloning, similarly to the strategy adopted in \citeappen{starcraf19}. Our overall learning objective during policy optimization is a linear combination of policy gradient, value function regression loss, and behavior cloning.

In the meantime but optionally, as we have already limited the amount of shift, one can also incorporate state distribution correction into the policy gradient estimate. In our implementation, the shift is further accounted for through reweighing a state with the production of all previous IS ratios $w_t = \prod_{k=0}^{t-1}\frac{\pi(a_k|s_k)}{\pi_b(a_k|s_k)}$. Similar to weighted importance sampling during off-policy policy evaluation, we can normalize $w_t$ within each training batch to reduce variance. And the normalizer is regularized to prevent all weights being annihilated under normalization. In theory, any regularization that penalizes minuscule values in the normalized weights could confer such function, for instance, imposing soft-maximum over $\log w_t$ \citeappen{GPSlevine13}, or minimizing the effective sample size \citeappen{Doerr2019}.

\section{Off-Policy Evaluations}
\label{section:supp_OPE}

\begin{figure}[!htp]
	\centerline{
			\includegraphics[width=1.\textwidth]{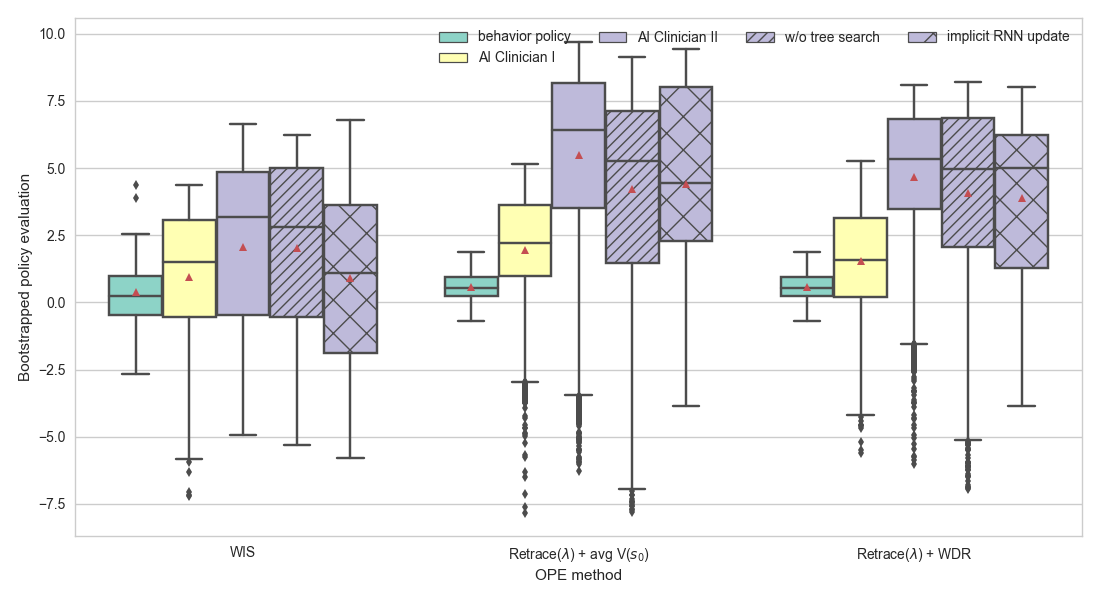}\\
	}
\caption{Ablations on the test set under multiple OPE methods.}
\label{fig:rl_results}
\end{figure}

We evaluate our learned policy in terms of off-policy evaluation (OPE), for which we choose three approaches with varying balances between evaluation bias and variance: weighted importance sampling\footnote{Each per-trajectory IS ratio is clipped to $[1e-30, 1e10]$.}, averaging the state value across all encountered initial states upon estimating a state value function using Retrace$(\lambda)$ \citeappen{Munos2016retrace}, and weighted doubly-robust (WDR) \citeappen{56Thomas2016DataEfficientOP} with the same approximated state value function. The results are shown concurrently with two of ablation studies in Figure \ref{fig:rl_results}, for explicitly learning state representation and the tree search respectively. The results demonstrate that AI Clinician II is consistently more optimal than both behavior policy and AI Clinician I, and that the two algorithmic building blocks are indispensable.

\end{document}